\documentclass[fleqn,10pt]{JLA_article} 

\usepackage[english]{babel} 
\usepackage{longtable}


\newcommand{\STEPS}{the Guided AI Tutor\xspace}
\newcommand{\STEPSfull}{the Guided AI Tutor (name withheld for review)\xspace}

\usepackage{hyperref} 
\usepackage{array}
\usepackage{xspace}
\hypersetup{hidelinks,colorlinks,breaklinks=true,urlcolor=color2,citecolor=color1,linkcolor=color1,bookmarksopen=false,pdftitle={Title},pdfauthor={Author}}

\PaperTitle{Examining Variation in How Guided AI Tutors Resolve Student Impasses} 

\Authors{Bakhtawar Ahtisham\textsuperscript{1}, Kirk Vanacore\textsuperscript{1}, Alessandra Napoli\textsuperscript{2}, Josh Arens\textsuperscript{2}, Ksenia Ionova\textsuperscript{1}, Clayton Cohn\textsuperscript{1}, Shima Salehi\textsuperscript{2}, Rene Kizilcec\textsuperscript{1}} 
\affiliation{\textsuperscript{1}\textit{Cornell University, Ithaca, NY, United States}} 
\affiliation{\textsuperscript{2}\textit{Stanford University, Stanford, CA, United States}} 

\Submitted{09/28/2026}
\Accepted{-}
\Published{-}

\Volume{1}
\Number{1}
\Pages{1---10}
\Doi{xxx-xxx-xxx}
\Notesname{Notes for Practice} 
\note{Pedagogical guardrails in LLM tutors (e.g., ``do not give the answer'') change \emph{what} the tutor withholds, but do not by themselves specify \emph{how} it should adapt when a student stays stuck.}
\note{In authentic dialogues with a guided tutor, questioning helped most at the first sign of an impasse; its benefit shrank with every additional turn the student remained stuck, whereas directly addressing the student's error became relatively more helpful.}
\note{Students who abandoned sessions early were caught in repeated concept-elicitation loops before ever reaching problem execution.}
\note{Learning analytics systems can track impasse depth and type in tutoring dialogues in real time to flag students who remain stuck, and support graduated escalation, moving from questioning to targeted error feedback and, when needed, worked sub-steps.}
\Abstract{When a student is stuck, a tutor faces the assistance dilemma: help given too early can hinder productive struggle, while help withheld too long leaves the student in a frustrating, persistent impasse (i.e., wheel spinning). Generative AI tutors increasingly use guardrails restricting answer-giving, yet little is known about how such tutors behave once an impasse persists. We analyze 20,462 student turns from 1,260 authentic sessions with a guided LLM chemistry tutor, identifying 6,630 impasse turns of three major types: conceptual errors, expressed uncertainty, or help-seeking. We then used these impasses to simulate three tutoring conditions to study variation in AI tutor guidance through impasses: baseline, no-direct-answer, and guided tutor. For a sample of 150 impasses, prompt specificity changed pedagogy: a baseline tutor provided the answer directly in 50.7\% of responses, a no-direct-answer tutor asked a follow-up question every time, and the guided tutor responded in a wide variety of ways depending on the context. We then analyzed impasse trajectories in authentic interactions, finding that each additional impasse turn lowered the odds of next-turn recovery by 12.7\% ($\mathrm{AOR}=0.873$, $p<.001$), and early dropouts were caught in recursive concept elicitation before reaching execution. The benefit of questioning decayed as impasses persisted (scripted question $\times$ depth $\mathrm{AOR}=0.78$; follow-up $\times$ depth $\mathrm{AOR}=0.83$), whereas addressing the student's error grew more beneficial ($\mathrm{AOR}=1.14$); after a failed scripted question, repeating it was followed by recovery in 28.1\% of cases, compared with 39.8\% when the tutor addressed the error instead. For learning analytics, these findings identify impasse depth and type as observable, turn-level dialogue signals that analytics can use to trigger graduated, state-sensitive assistance in real time.\\ }

\begin{document}
\pagestyle{plain}
\fancyhead{}
\renewcommand{\headrulewidth}{0pt}
\flushbottom 

\maketitle 


\thispagestyle{plain} 


\section{Introduction} 

\addcontentsline{toc}{section}{Introduction} 
Being stuck is a normal, and often productive, part of learning to solve problems. Theories of impasse-driven learning describe these moments as \textit{impasses}: states in which a learner's current knowledge is insufficient to produce a correct next step and, precisely because of this failure, new knowledge may be constructed \cite{vanlehn1988toward,vanlehn2003why}. However, impasses are productive only when they are resolved. When students cannot repair them, productive struggle becomes unproductive failure \cite{kapur2016examining} or \emph{wheel-spinning}---in which students repeatedly fail to master a skill despite extended practice \cite{beck2013wheel,gong2015towards,kai2018decision}---and may lead to disengagement. Deciding when and how much to help is therefore a central problem for any tutor, a problem \citeA{koedinger2007exploring} named the \emph{assistance dilemma}: both too much and too little assistance can impair learning, and the optimal level depends on the learner's state.
 
Large language models (LLMs) have made it cheap to deploy conversational tutors at scale, and early evidence is mixed. Well-designed AI tutors can produce substantial learning gains \cite{kestin2025ai,pardos2024chatgpt}, but unrestricted access to a general-purpose model can improve practice performance while harming later unassisted learning \cite{bastani2025generative}. The prevailing response is to add \emph{pedagogical guardrails}: system prompts that forbid giving away answers and ask the model to guide students with questions and hints \cite{liffiton2023codehelp,bastani2025generative,jurenka2024towards}. Such guardrails address the risks of excessive assistance but provide less guidance about what a tutor should do when its initial support is insufficient, and the student remains stuck.
 
This distinction matters because LLM tutors are typically evaluated one response at a time, for example by rating whether a single turn avoids revealing the answer or identifies the student's mistake \cite{tack2022ai,maurya2025unifying}. Such evaluations cannot reveal whether a tutor \emph{adapts} across turns as evidence accumulates that its assistance is failing. Human tutoring research, by contrast, has long emphasized contingency: effective tutors increase the specificity of their help after a failure and reduce it after success \cite{wood1976role,wood1999help}. Whether guardrailed LLM tutors behave contingently, and whether contingency matters for students in authentic use, is an open empirical question for learning analytics.
 
We address this question with log data from a constrained LLM tutor that supports chemistry students in planning solutions to multi-step problems, organized around expert problem-solving practices \cite{price2021detailed,burkholder2020template,schwartz2024template}. We combine a controlled replay of student impasses through tutors with different levels of constraint with sequential and multivariable analyses of the tutor's authentic dialogues. We ask:
 
\begin{itemize}
    \item[\textbf{RQ1.}] How do student impasses manifest across problem-solving practices in tutoring dialogues?
    \item[\textbf{RQ2.}] How does the degree of prompt-based guidance change the way an AI tutor responds to student impasses?
    \item[\textbf{RQ3.}] How do students who complete the session differ from those who drop out in how they progress through problem-solving practices?
    \item[\textbf{RQ4.}] Which tutor moves are associated with recovery from student impasses?
\end{itemize}

This paper makes three contributions to learning analytics. \emph{Conceptually}, it extends the evaluation of guardrailed LLM tutors beyond single-turn compliance by examining how tutor support adapts across persistent student impasses. \emph{Empirically}, it shows, in 1,260 authentic sessions, that the benefit of the tutor's dominant move of questioning decays as an impasse persists, while error-directed feedback grows relatively more useful, and that early attrition is concentrated in concept-elicitation loops. \emph{Methodologically}, it offers a reusable operationalization of impasse episodes, depth, and recovery from codebook-annotated dialogue logs. Together, these findings motivate \emph{graduated assistance}: tutor policies that condition the type and degree of support on how long a student has been stuck.

\section{Literature Review}
\subsection{Impasses, Productive Struggle, and the Assistance Dilemma}
\citeA{vanlehn1988toward} proposed that learning is driven by impasses: when a learner's procedures cannot produce the next step, they must repair the gap, and the repair may be generalized into new knowledge. Studies of problem solving \cite{vanlehn1999rule} and human tutoring \cite{vanlehn2003why} found that learning events concentrate at impasses, and that tutor explanations are most effective when they follow a student's impasse rather than precede it. Related work on affect shows that confusion, the affective signature of an impasse, can benefit learning when it is resolved, but becomes harmful when it persists into frustration or boredom \cite{dmello2014confusion,baker2010better}.
 
Research on productive failure extends this idea to instructional design: allowing students to struggle before receiving instruction can deepen conceptual understanding \cite{kapur2008productive}, but the benefits depend on how that struggle is supported \cite{kapur2016examining}. This creates a practical version of the assistance dilemma \cite{koedinger2007exploring}: support must be sufficient to help learners progress without eliminating opportunities for productive problem solving. Our work examines this tradeoff turn by turn, asking how the value of different forms of assistance changes as an impasse persists.

\subsection{Help-Seeking, Contingent Tutoring, and Graduated Assistance}
Classic accounts of scaffolding describe tutors who adjust their support to the learner's success or failure \cite{wood1976role}. \citeA{wood1999help} formalized this as \emph{contingent tutoring}: increase the specificity of help after a failure and reduce it after success. Studies of naturalistic human tutoring show that tutors rely heavily on questioning, feedback, and collaborative repair of errors \cite{graesser1995collaborative}, and that eliciting student constructions can be as effective as tutor explanations \cite{chi2001learning}. Intelligent tutoring systems implement a version of contingency through hint sequences that move from general prompts to bottom-out hints \cite{vanlehn2011relative}, and a substantial literature models when and how students seek such help \cite{aleven2003help,aleven2006toward,price2017hint}. Help-seeking is a meaningful signal in its own right: explicit requests indicate metacognitive awareness of an impasse, whereas hedged answers signal partial knowledge. We therefore treat conceptual errors, expressed uncertainty, and help requests as distinct manifestations of an impasse rather than a single failure state.
 
\subsection{Generative AI Tutors and Pedagogical Guardrails}
LLM tutors differ from rule-based tutoring systems in that their pedagogical behavior is specified largely through natural-language instructions rather than an explicit tutoring model. Field and laboratory studies report both promise and risk \cite{yan2024generative}. ChatGPT-generated hints produced learning gains comparable to human-authored hints \cite{pardos2024chatgpt}, and a research-designed AI tutor outperformed in-class active learning in a university physics course \cite{kestin2025ai}. However, unrestricted access to GPT-4 harmed high-school students' subsequent unassisted performance, while a version prompted to give hints rather than answers largely removed this harm \cite{bastani2025generative}. Similarly, how students use LLMs, as a substitute for or complement to their own work, shapes whether they learn \cite{lehmann2024ai}. Indeed, surveys of STEM college students show a near even split between these two practices, with 54\% of students using AI as a scaffold for their own reasoning and 46\% as a shortcut to answers \cite{wang2025scaffold}.
 
In response, researchers and developers have built guardrailed tutors that withhold solutions \cite{liffiton2023codehelp,sheese2024patterns}, fine-tuned models for pedagogy \cite{jurenka2024towards,macina2023mathdial,scarlatos2025training}, and developed methods for diagnosing and remediating specific student errors \cite{daheim2024stepwise}. Evaluation frameworks rate tutor responses along pedagogical dimensions such as mistake identification, answer revelation, and actionability \cite{tack2022ai,maurya2025unifying}. Human--AI approaches such as Tutor CoPilot show that LLM suggestions can shift human tutors toward more guiding questions and fewer revealed answers \cite{wang2024tutor,thomas2024improving}. Most of this work evaluates responses in isolation or uses simulated students. Fewer studies analyze authentic student--LLM dialogues over time, and, to our knowledge, none examines whether a guardrailed tutor's assistance adapts as a student's impasse persists.
 
\subsection{Temporal Analytics of Tutoring Dialogue}
Learning analytics has increasingly argued that the order and timing of events, not only their frequency, carry information about learning processes \cite{reimann2009time,chen2018critical,knight2017time}. Sequential methods such as transition matrices and lag analysis have been used to characterize tutoring and collaborative dialogue \cite{bakeman1995analyzing,borchers2024combining}, and recent work applies LLMs to trace knowledge directly from tutor--student dialogue \cite{scarlatos2025exploring}. Discrete-time event models provide a natural framework for asking how the probability of an event, here, resolving an impasse, changes with elapsed time and with time-varying covariates such as the tutor's most recent move \cite{singer1993its}. We draw on these traditions to model impasse episodes as turn-level processes nested within sessions.

\vspace{\baselineskip}
Taken together, prior work points to a common problem: assistance is often evaluated as a property of a single tutor response or as a fixed instructional policy, rather than as something that should adapt as evidence accumulates that a student remains stuck. Research on impasses and contingent tutoring suggests that support should become more specific after unsuccessful attempts, while work on LLM guardrails shows how prompts can constrain tutor behavior but says much less about whether that behavior changes appropriately across turns. Learning analytics provides methods for modeling sequential interaction processes, yet these methods have rarely been used to examine how the value of specific LLM tutor moves changes within persistent impasses. We address this gap by combining a controlled comparison of tutors with different levels of prompt guidance and turn-level analysis of authentic student--LLM dialogues, using impasse type, depth, tutor moves, and next-turn recovery to examine how assistance changes as students remain stuck.
\section{Methods}

\subsection{Guided AI Tutor and Context}\label{sec:steps}
The dialogue data used in this study were collected using \STEPSfull, an LLM-based tutor (GPT-4o via the OpenAI API) that guides students to plan a solution to a multistep problem without giving them the solution \cite{anon2026development}. The tutor's behavior is controlled entirely by its prompt, which is based on validated problem-solving templates for chemistry \cite{schwartz2024template} and physics \cite{burkholder2020template}. The prompt covers five parts: (1) a \emph{role} as a problem-solving coach; (2) \emph{objectives} to help students build their own plan, reflect metacognitively, and stay engaged; (3) \emph{behavioral guidelines}: never solve the problem, ask one question at a time, keep responses to three sentences or fewer, do not praise incorrect answers, and after an incorrect or incomplete answer give ``increasingly specific, targeted hints''; (4) the \emph{problem and four scripted questions}, each paired with its correct answer so the tutor can check responses against a reference; and (5) \emph{response-handling rules} for answer requests and off-topic turns. Two features matter for our study. First, the scripted sequence determines which practice each exchange targets, so the tutor, not the student, decides when the dialogue moves on. Second, escalation after errors is requested only as a natural-language instruction. This deployed configuration is called ``Guided AI Tutor'' in the analyses addressing RQ2. The tutor's planning sequence is organized around \emph{problem-solving practices}, the decisions experts make when solving authentic problems \cite{price2021detailed}, adapted into a planning template for chemistry \cite{schwartz2024template}. We code eight practices: problem identification, relevant concepts, similar problems, information needed, assumptions, strategy, execution, and answer checking (defined in Table~\ref{tab:codebook}).

\paragraph{Data and ethics.} Data were collected in an introductory general chemistry course at a large research university in the United States, where students used the tutor through a stand-alone web application to plan solutions to 8 assigned chemistry problems. The model version and configuration were held fixed throughout data collection. The corpus comprises 1,260 sessions (20,462 student turns; mean 16.2 per session). This study is a secondary analysis of existing tutoring logs; all transcripts were de-identified before analysis and contained no personally identifiable information. The study was determined exempt by the authors' Institutional Review Board [protocol number withheld for review].

\subsection{Codebook Development}\label{sec:codebook}
The codebook was first developed deductively for a physics implementation of the tutor, building on existing frameworks for evaluating AI tutors \cite{tack2022ai,petukhova2025intentmattersenhancingai,maurya2025unifying,pauzi2025automating}. It initially described tutor moves (e.g., asking questions, giving feedback); codes for student behavior (e.g., answering, help-seeking) and for the correctness of student and tutor statements were added to follow the conversational flow. For the chemistry implementation, the problem-solving practices of the planning template \cite{price2021detailed,schwartz2024template} were added as deductive codes, alongside inductive codes that emerged from the data. Two researchers refined the codebook iteratively, coding the same two to four conversations, discussing disagreements, and splitting or merging codes until definitions stabilized. The full codebook also includes categories not analyzed here (e.g., tutor correctness identification, answer leakage) and is available as supplementary material; the codes used in this study are listed in Table~\ref{tab:codebook}.

\begingroup
\small
\setlength{\tabcolsep}{4pt}
\renewcommand{\arraystretch}{1.15}
\begin{longtable}{>{\raggedright\arraybackslash}p{0.17\textwidth} >{\raggedright\arraybackslash}p{0.27\textwidth} >{\raggedright\arraybackslash}p{0.50\textwidth}}
\caption{Codebook categories and codes used in the analyses. Student-input and correctness codes define impasse types and recovery (Section~\ref{sec:impasse}); tutor-output codes are the moves analyzed in RQ2 and RQ4.}
\label{tab:codebook}\\
\toprule
\textbf{Category} & \textbf{Code} & \textbf{Definition} \\
\midrule
\endfirsthead
\toprule
\textbf{Category} & \textbf{Code} & \textbf{Definition} \\
\midrule
\endhead
\bottomrule
\endlastfoot
\textbf{Student input} & Direct answer or explanation & Response to a question and/or explanation of thinking. \\
 & Answer with uncertainty & Response to a question expressed with doubt or uncertainty. \\
 & Specific question & Question seeking specific information or asking about a specific step. \\
 & General help-seeking & Expresses confusion or general uncertainty without referencing specific information or a step. \\
\midrule
\textbf{Correctness} & Completely correct & Fully accurate and complete, with no errors or missing key information (defines recovery). \\
 & Partially incorrect & Some components are correct, but at least one key idea, step, or piece of information is incorrect or missing. \\
 & Completely incorrect & Contains fundamental factual errors that make the reasoning or result invalid. \\
 & Not specific enough & Broadly correct but too vague to determine whether the student can apply it in context. \\
\midrule
\textbf{Tutor output} & Scripted question & Asks a question explicitly included in the prompt's scripted sequence. \\
 & Follow-up question & Asks additional questions to clarify, extend, or probe the student's thinking. \\
 & Check for understanding & Explicitly asks whether the student understands a concept, explanation, or step. \\
 & Direct answer & Directly responds to the student's explicit question or request for information. \\
 & Confirm / elaborate & Acknowledges a correct response and may provide additional explanation or context. \\
 & Address incorrect answer & Responds to incorrect or partially incorrect work by correcting, guiding, or explaining. \\
 & Partial answer / hint & Gives partial solution information or scaffolding without fully solving the problem. \\
 & Praise / encouragement & Gives supportive, motivating, or affirming feedback. \\
 & Remind exercise purpose & Restates the goals or intended learning purpose of the activity. \\
\midrule
\textbf{Problem-solving practice} & Problem identification & Identifies or restates the type, goal, or key components of the problem. \\
 & Relevant concepts & Identifies concepts, principles, or skills needed to solve the problem, separate from applying them. \\
 & Similar problems & Identifies and/or compares a similar problem. \\
 & Information needed & Identifies or gathers specific information, values, equations, or reference data, separate from using them. \\
 & Assumptions & Makes assumptions or simplifications that help define or solve the problem. \\
 & Strategy & Plans the steps or sequence of actions needed to solve the problem. \\
 & Execution & Carries out solution steps or calculations to obtain the result. \\
 & Answer checking & Checks or evaluates whether a solution, step, or answer is correct. \\
\end{longtable}
\endgroup

\subsection{Ground Truth Construction and Annotation}\label{sec:annotation}

We annotated the corpus using a human-validated LLM annotation pipeline built around \STEPS\ dialogue codebook (Table~\ref{tab:codebook}). First, two annotators with chemistry domain expertise independently coded 166 utterances from six randomly selected sessions on all codebook dimensions with an initial inter-rater reliability of 83\%; remaining disagreements were resolved through discussion, and code definitions were refined where they proved ambiguous to provide a single agreed-upon ground truth for an LLM annotator. We then applied an LLM annotator (Gemini 3.1 Pro), prompted with the refined codebook definitions and examples, to 230 utterances from 10 further randomly sampled sessions. We refined the prompt until reaching high agreement between the LLM and human annotations (mean $\kappa = 0.86$). Because prior work has found that LLMs can match or exceed human annotators on some text-annotation tasks \cite{gilardi2023chatgpt}, we did not assume that disagreements necessarily reflected errors by the LLM. Instead, two human annotators independently reviewed every AI-assigned label, either confirming it or replacing it with the appropriate code. These adjudicated labels served as the gold standard.

To scale the annotation across the entire corpus, we used the final prompt on this gold-standard set, which yielded substantial agreement~\cite{cohen1960coefficient,landis1977measurement} across the codebook dimensions used in our analyses: student input type (mean $\kappa=.84$), tutor moves (mean $\kappa=.79$), and problem-solving practice (mean $\kappa=.80$). For correctness, the code that defines recovery, \emph{completely correct}, was identified reliably ($\kappa=.81$). Rare or unreliable codes (e.g., \emph{ignore student question}, \emph{unintelligible}) were excluded, and the validated prompt was then applied to the full corpus of 1,260 sessions. The same validated prompt was used to code the tutor responses in the controlled comparison (Section~\ref{sec:rq2methods}).

\subsection{Operationalizing Student Impasses}\label{sec:impasse}
Following impasse-driven learning theory \cite{vanlehn1988toward,vanlehn2003why}, we treat an impasse as a state in which the student's current knowledge is insufficient to produce a correct next step, visible in dialogue as an error, hesitation, or a request for help. We define four constructs over the annotated logs:
 
\begin{enumerate}[noitemsep, topsep=0pt]
    \item \textbf{Impasse turn.} A student turn coded as (a) a \emph{conceptual error}: a direct answer or explanation coded as completely incorrect, partially incorrect, or not specific enough; (b) \emph{expressed uncertainty}: an answer with explicit hedging (e.g., ``I think'', ``maybe'', ``is it...?''); or (c) \emph{help-seeking}: a general help request (``I don't understand'', ``idk'') or a specific question that proposes no answer. The three types are mutually exclusive by construction.
    \item \textbf{Impasse episode.} A maximal run of consecutive impasse turns within a session.
    \item \textbf{Impasse depth.} The position of a turn within its episode ($1,2,\dots,k$); depth 1 is the onset.
    \item \textbf{Recovery.} The student's next turn ($t+1$) is coded as a completely correct answer, which ends the episode. Episodes that end without a correct response (e.g., the session ends) are unresolved.
\end{enumerate}

Across the corpus we identified 6,630 impasse turns in 4,170 episodes within 1,204 sessions: 2,311 conceptual errors (34.9\%), 1,970 expressions of uncertainty (29.7\%), and 2,349 help requests (35.4\%). 
 \begin{table}[t]
\centering
\caption{Distribution of impasse turns across problem-solving practices and the share of each impasse type within a practice ($N=6{,}630$).}
\label{tab:impasse_distribution}
\small
\begin{tabular}{lrrrrr}
\toprule
\textbf{Practice} & \textbf{Turns} & \textbf{\%} & \textbf{Error \%} & \textbf{Uncert.\ \%} & \textbf{Help \%} \\
\midrule
Relevant concepts      & 3,465 & 52.3 & 27.8 & 14.1 & 58.1 \\
Strategy               & 1,197 & 18.1 & 44.8 & 38.8 & 16.5 \\
Execution/solution     &   961 & 14.5 & 50.7 & 48.3 &  1.0 \\
Information needed     &   508 &  7.7 & 46.3 & 35.6 & 18.1 \\
Answer checking        &   345 &  5.2 &  3.8 & 95.1 &  1.2 \\
Problem identification &   110 &  1.7 & 51.8 & 25.5 & 22.7 \\
Assumptions            &    31 &  0.5 & -- & -- & -- \\
Similar problems       &    13 &  0.2 & -- & -- & -- \\
\bottomrule
\end{tabular}
\end{table}
\subsection{Analyses}
\paragraph{RQ1: Manifestation and persistence.}
To evaluate when specific impasses occur, we cross-tabulated impasse type by problem-solving practice and tested independence with a Pearson $\chi^2$ test (Cram\'er's $V$ as effect size). We then computed next-turn recovery, whether the impasse is recovered on the subsequent turn, by depth and by impasse type, with 95\% confidence intervals from a session-level cluster bootstrap. This allows us to assess how persistent each impasse type is, and how the likelihood of recovery changes as impasse depth increases. 

\paragraph{RQ2: Controlled comparison of tutor conditions.} \label{sec:rq2methods}
To evaluate how prompt-based guidance influences an AI tutor's responses to student impasses, we ran a small simulation study. First, we drew a stratified random sample of 150 impasse turns ($n=50$ per impasse type). For each of the sampled impasse turn, we prompted an AI model with three experimental tutoring conditions: \emph{Baseline AI Tutor}, \emph{No-Direct-Answer Baseline AI Tutor}, \emph{Guided AI Tutor}. Each prompt included the problem on which the student had the impasse, preceding dialogue, and student utterance. The \emph{Baseline AI Tutor} was instructed only to respond as a tutor. The \emph{No-Direct-Answer Baseline AI Tutor} was additionally instructed to guide the student toward the solution rather than give the answer. The \emph{Guided AI Tutor} is the response the deployed system actually produced, governed by a detailed protocol: elicit a structured plan through sequential questions organized around expert problem-solving practices, ask one question at a time, avoid supplying correct answers after incorrect or incomplete responses, give increasingly specific hints when needed, keep responses concise, and encourage without praising incorrect answers. 
All comparison conditions were generated with GPT 5.6, and all responses were coded with the same multi-label set of tutor moves (scripted question, follow-up question, check for understanding, direct answer, confirm/elaborate, address incorrect answer, partial answer or hint, praise/encouragement, remind of exercise purpose). We compare conditions on move prevalence and on \emph{response diversity}, the number of distinct move combinations across the 150 responses.
We tested whether the prevalence of each move differed across the three conditions using Pearson $\chi^2$ tests (Bonferroni-corrected across the nine moves; Cram\'er's $V$ as effect size), and quantified response diversity as the number of distinct move combinations and their Shannon entropy.

\paragraph{RQ3: Session outcomes and practice transitions.}
Because a short session may reflect either fast success or early abandonment, we classified sessions using both length and progress. \emph{Early dropouts} ($N=252$, 20.0\%) ended with fewer than 10 student turns (corpus mean 16.2) without ever reaching the execution or answer-checking practices. \emph{Completers} ($N=635$, 50.4\%) reached execution or answer checking, including 68 short sessions ($<10$ turns) that did so. The remaining 374 sessions (29.7\%) exceeded 10 turns without reaching execution and were excluded to keep the contrast unambiguous. For each group, we estimated a first-order transition matrix over the six core practices, $P(\mathrm{PSP}_{t+1}=j \mid \mathrm{PSP}_t=i) = n_{ij}/\sum_k n_{ik}$, and the difference $\Delta\mathbf{P} = \mathbf{P}_{\text{Completers}} - \mathbf{P}_{\text{Dropouts}}$. Two rare practices (similar problems, $N=13$; assumptions, $N=31$) were excluded. Because dropouts are defined by never reaching execution, we interpret only transitions among the four pre-execution practices.
 
\paragraph{RQ4: Tutor moves and recovery.}
We fit turn-level logistic regressions predicting recovery at $t+1$ from the tutor's response to impasse turn $t$, using all impasse turns with an observed next student turn ($N=6{,}163$ turns, 3,917 episodes, $K=1{,}151$ sessions). Standard errors are cluster-robust (Huber--White) by session. All models control for impasse depth (the turn's position within its episode), impasse type (reference: conceptual error), and problem-solving practice (reference: relevant concepts). Tutor-move indicators are not mutually exclusive; most tutor turns combine several moves.
 
\textit{Model 1 (current and preceding moves)} includes indicators for seven tutor moves in the response to turn $t$ ($\mathbf{M}_t$), indicators for five moves in the tutor's response to the preceding impasse turn of the same episode ($\mathbf{M}_{t-1}$; zero at onset), and an indicator for repeating the same dominant move on consecutive turns:
\begin{equation}
\operatorname{logit} P(Y_{t+1}=1) = \beta_0 + \beta_d\,\mathrm{Depth}_t + \boldsymbol{\gamma}^\top \mathbf{X}_t + \boldsymbol{\beta}^\top \mathbf{M}_t + \boldsymbol{\eta}^\top \mathbf{M}_{t-1} + \beta_r\,\mathrm{Repeat}_t,
\end{equation}
where $\mathbf{X}_t$ holds impasse type and practice.
 
\textit{Model 2 (dominant two-turn sequences)} asks whether specific pairings of consecutive tutor moves matter, which Model~1 cannot show because it estimates the current and preceding moves as separate, additive effects; for example, Model~2 can compare repeating a scripted question after it has failed with switching to addressing the student's error. It assigns each tutor response a dominant move (priority: address incorrect $>$ check for understanding $>$ follow-up question $>$ scripted question $>$ partial answer $>$ confirm/elaborate $>$ direct answer) and replaces the move indicators with the sequence $M_{t-1}\rightarrow M_t$ within the episode (sequences with $\geq 25$ occurrences; others pooled), with \textit{Onset $\rightarrow$ Scripted question} as the reference. Because most tutor responses combine several moves, the priority order ranks moves from most to least diagnostic of the tutor's strategy, so that supportive moves that accompany almost every response (confirmation, praise) do not mask the move that directs the dialogue. We verify that this choice does not drive the results (Section~\ref{sec:robust}).
 
\textit{Model 3 (depth $\times$ move)} adds to the current-move specification interactions between depth and five moves central to the assistance dilemma (scripted question, follow-up question, address incorrect answer, partial answer, direct answer), testing whether a move's association with recovery changes as an impasse persists:
\begin{equation}
\operatorname{logit} P(Y_{t+1}=1) = \beta_0 + \beta_d\,\mathrm{Depth}_t + \boldsymbol{\gamma}^\top \mathbf{X}_t + \boldsymbol{\beta}^\top \mathbf{M}_t + \boldsymbol{\delta}^\top (\mathrm{Depth}_t \times \mathbf{M}_t).
\end{equation}
This is a discrete-time event model of episode resolution \cite{singer1993its}. We report adjusted odds ratios (AOR) with 95\% confidence intervals and, for Model 3, average predicted recovery with each move present versus absent at depths 1--4. As robustness checks (Section~\ref{sec:robust}), we re-estimated Model 3 with generalized estimating equations (exchangeable correlation) \cite{liang1986longitudinal} and Model 1 with depth as a categorical variable (1--5, 6+).

\section{Results}
\begin{figure}
    \centering
    \includegraphics[width=1\linewidth]{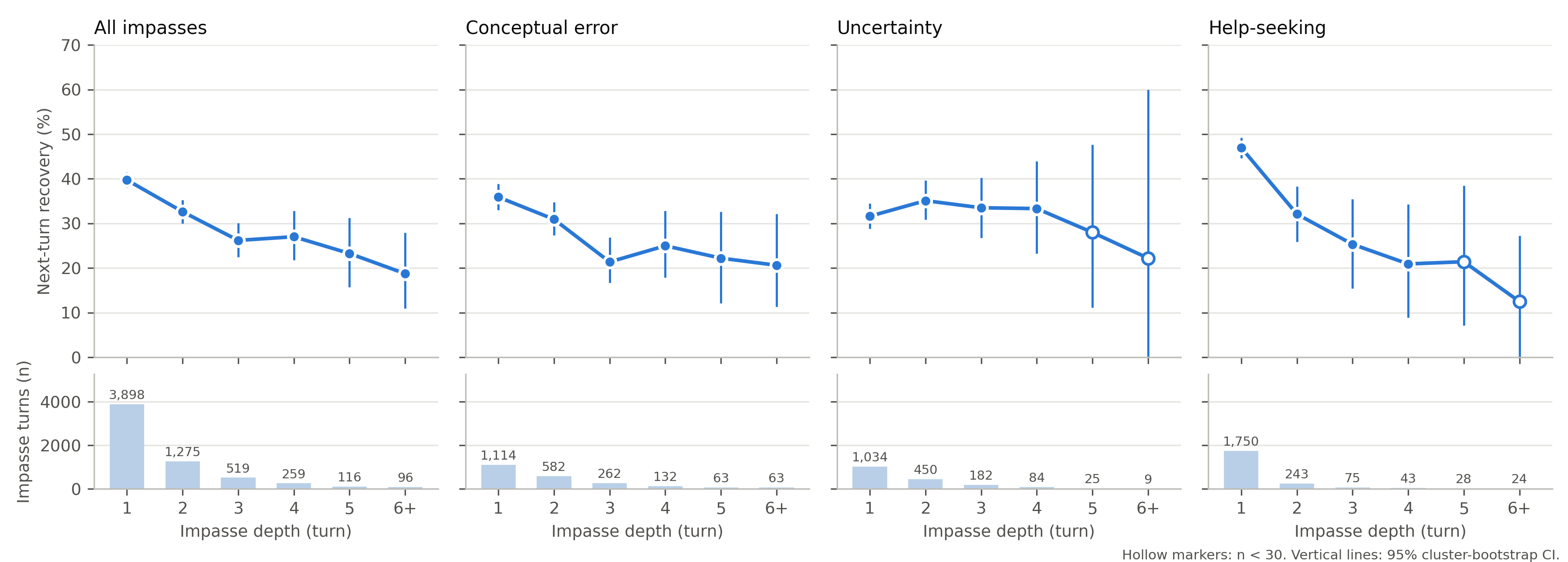}
    \caption{Next-turn recovery by impasse depth, for all impasses and by impasse type. Error bars show 95\% confidence intervals from a session-level cluster bootstrap (2,000 resamples); the number of impasse turns at each depth appears in the bottom panel. Hollow markers indicate fewer than 30 turns.}
    \label{fig:depth_type}
\end{figure}
\begin{table}
\centering
\small
\caption{Prevalence of tutor moves across the three tutoring conditions for the same 150 impasses, with tests of between-conditions differences. Cells give the percentage of each condition's responses that contained the move; because a single response could contain several moves, columns do not sum to 100\%. \emph{Response diversity} counts the distinct combinations of moves each condition produced and their Shannon entropy (higher values indicate more varied responses).}
\label{tab:rq1_tests}
\begin{tabular}{lccccc}
\toprule
 & \multicolumn{3}{c}{\textbf{\% of responses}} & & \\
\cmidrule(lr){2-4}
\textbf{Tutor move} & \textbf{Baseline} & \textbf{No-Direct-Answer} & \textbf{Guided} & $\boldsymbol{\chi^2(2)}$ & \textbf{Cram\'er's $V$} \\
\midrule
Direct answer            & 50.7 & 0.0   & 19.3 & 109.6 & .49 \\
Partial answer / hint    & 43.3 & 0.0   & 39.3 &  86.2 & .44 \\
Follow-up question       & 62.7 & 100.0 & 35.3 & 140.9 & .56 \\
Address incorrect answer & 37.3 & 33.3  & 16.0 &  18.8 & .20 \\
Confirm / elaborate      & 12.0 & 7.3   & 42.0 &  65.3 & .38 \\
Check understanding      & 0.0  & 0.0   & 23.3 &  75.9 & .41 \\
Scripted question        & 0.0  & 0.0   & 31.3 & 105.0 & .48 \\
Praise / encouragement   & 0.0  & 0.0   & 35.3 & 120.2 & .52 \\
Remind exercise purpose  & 0.0  & 0.0   & 12.0 &  37.5 & .29 \\
\midrule
\multicolumn{6}{l}{\textit{Response diversity}} \\
Distinct move combinations & 14   & 3    & \textbf{51}   & & \\
Shannon entropy (bits)     & 3.11 & 1.25 & \textbf{5.13} & & \\
\bottomrule
\multicolumn{6}{l}{\footnotesize All $\chi^2$ tests significant at Bonferroni-corrected $p<.001$ (nine tests).}\\
\end{tabular}
\end{table}

\subsection{RQ1: Where Impasses Occur and How Recovery Declines}
Impasses were concentrated in the planning phases of problem solving (Table~\ref{tab:impasse_distribution}): 52.3\% occurred while identifying relevant concepts, 18.1\% during strategy development, and 14.5\% during execution. These are counts of where impasses were observed, not practice-specific rates, since practices differ in exposure. Impasse type depended strongly on practice ($\chi^2(10)=2313.9$, $p<.001$, Cram\'er's $V=.42$). Help requests made up most of the impasses in relevant-concepts phase (58.1\%), conceptual errors and uncertainty were roughly balanced during strategy and execution (45--51\% errors), and nearly all answer-checking impasses were expressions of uncertainty (95.1\%).

Across all impasse types, next-turn recovery fell steadily as impasses persisted, from 39.7\% at onset to 32.6\% at depth 2, 26.2\% at depth 3, and 18.8\% at depth 6 or more, less than half the onset rate (Figure~\ref{fig:depth_type}). However, trajectories differed by type. Help-seeking was initially the most recoverable state (47.0\% at onset) but declined most sharply once initial assistance had failed (25.3\% at depth 3; 12.5\% at depth $\geq$6). Conceptual errors dropped to about 21--25\% by depth 3 and remained there (20.6\% at depth $\geq$6), whereas uncertainty remained comparatively recoverable through depth 4 (31.6\% at onset; 33.3\% at depth 4), beyond which few uncertainty episodes persisted.
 \begin{figure}[t]
     \centering
     \includegraphics[width=1\linewidth]{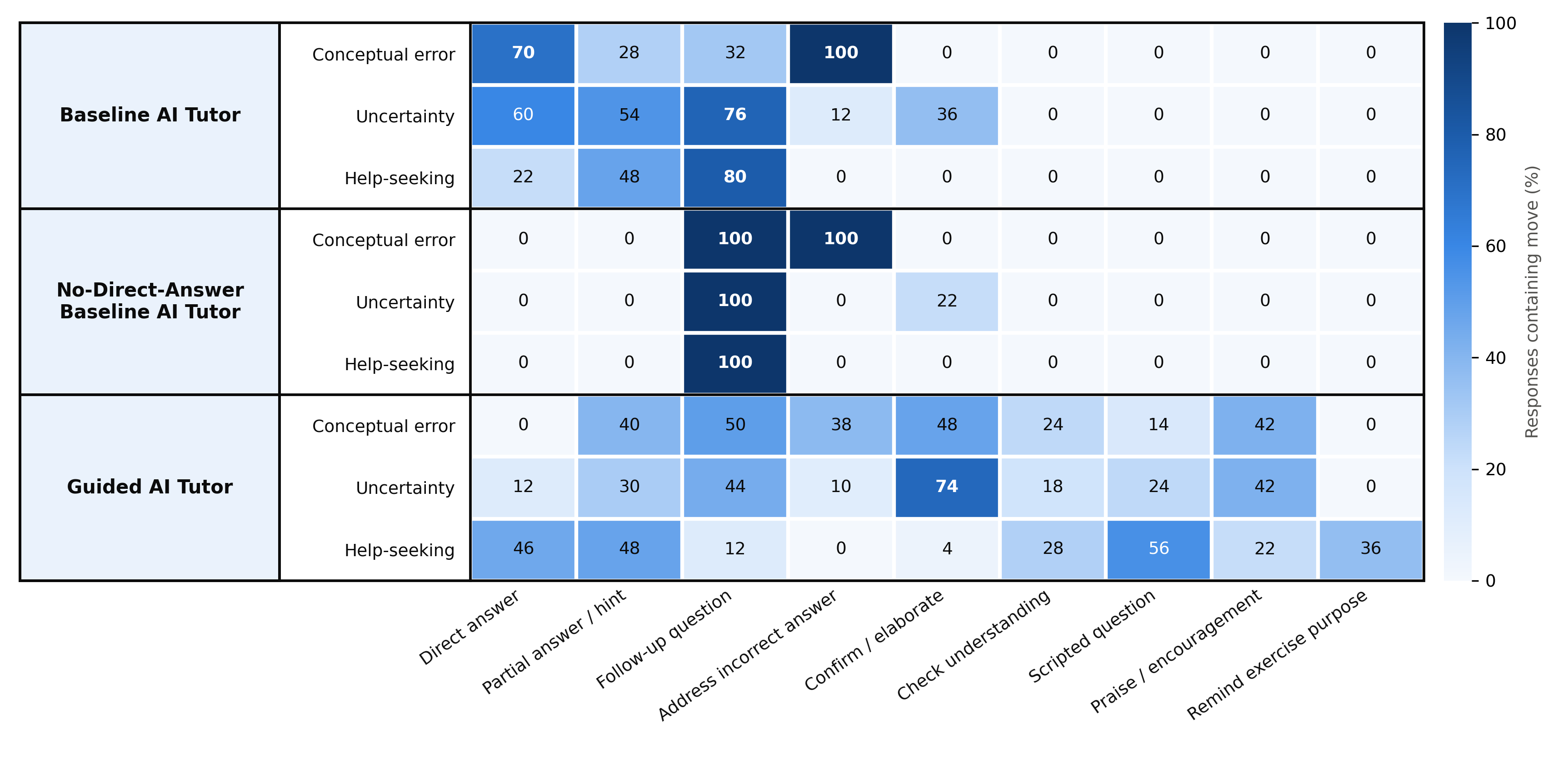}
     \caption{Pedagogical response profiles across tutor conditions and impasse types for the same 150 impasses ($n=50$ per type). Cells show the percentage of responses containing each move; responses can contain several moves, so rows need not sum to 100\%.}
     \label{fig:profiles}
 \end{figure}
 
\subsection{RQ2: Prompt Specificity Shapes Tutor Responses to Impasses}
The three conditions produced significantly different response profiles for the same 150 impasses (Figure~\ref{fig:profiles}). All nine moves differed significantly across conditions (all Bonferroni-corrected $p<.001$; Cram\'er's $V=.20$--$.56$), with the largest differences for follow-up questions ($V=.56$), praise ($V=.52$), direct answers ($V=.49$), and scripted questions ($V=.48$). Response diversity followed the same ordering: the Guided AI Tutor's move combinations had an entropy of 5.13 bits, compared with 3.11 for the Baseline AI Tutor and 1.25 for the No-Direct-Answer Baseline AI Tutor. These differences held when the analysis was restricted to the 104 impasses with an unambiguous problem context (all corrected $p<.05$).

The \textbf{Baseline AI Tutor} provided direct answers in 50.7\% of responses (vs.\ 19.3\% for the Guided AI Tutor and 0\% for the No-Direct-Answer Baseline AI Tutor). This tendency was stronger for conceptual errors (70\%), where it also addressed the incorrect answer in every case. It nevertheless asked follow-up questions in 62.7\% of responses. The \textbf{No-Direct-Answer Baseline AI Tutor} adopted a uniform questioning policy: it asked a follow-up question in 100\% of responses, never gave a direct or partial answer, and produced only 3 distinct move combinations. The \textbf{Guided AI Tutor} used the broadest repertoire, including moves absent from both alternatives: scripted questions (31.3\%), checks for understanding (23.3\%), praise or encouragement (35.3\%), and reminders of the exercise purpose (12.0\%). It also combined questioning with partial answers or hints (39.3\%) and confirmation or elaboration (42.0\%).
 
The Guided AI Tutor's responses also varied with the type of impasse. Help requests most often elicited scripted questions (56\%); conceptual errors elicited follow-up questions (50\%), confirmation or elaboration (48\%), hints (40\%), and error-directed feedback (38\%); and uncertainty most often elicited confirmation or elaboration (74\%). Table~\ref{tab:tutor_example} illustrates the contrast for one conceptual error: the Baseline AI Tutor produces the full solution procedure, the No-Direct-Answer Baseline AI Tutor redirects with questions, and the Guided AI Tutor names the misconception (the apparent mass is a weighted average and must lie between the two isotope masses) before asking a follow-up question.
 
In short, a limited prompt instructions not to give answers did not produce a nuanced pedagogy; it produced uniform questioning. A detailed protocol produced varied, impasse-sensitive responses. Constraining answer-giving is therefore not equivalent to specifying a pedagogical policy.


 \begin{table*}
\centering
\footnotesize
\setlength{\tabcolsep}{4pt}
\renewcommand{\arraystretch}{1.1}
\caption{Responses of the three tutor conditions to the same conceptual-error impasse. Coded tutor moves are shown in brackets and italicized.}
\label{tab:tutor_example}
\begin{tabular}{>{\raggedright\arraybackslash}p{0.15\textwidth} p{0.81\textwidth}}
\toprule
\textbf{Field} & \textbf{Content} \\
\midrule
\textbf{Problem Statement} & Naturally occurring cobalt consists of only one isotope, Co-59, whose relative atomic mass is 58.9332 amu. A synthetic radioactive isotope of cobalt, Co-60, has a relative atomic mass of 59.9338 amu and is used in radiation therapy for cancer. A doctor prepares a 1.7516 g sample of cobalt that is a mixture of these two isotopes and has an apparent atomic mass of 58.9901 amu. What number of Co-60 atoms are in this sample? \\
\textbf{Preceding tutor turn} & ``Great! Let's connect the chemistry knowledge to the problem. How do you think the different atomic masses of Co-59 and Co-60 relate to the reported atomic mass of the cobalt sample?'' \\
\textbf{Student impasse} & ``its greater than the both of them?'' \textit{(conceptual error: the apparent atomic mass is a weighted average, so it must lie \emph{between} the two isotope masses)} \\
\midrule
\textbf{Baseline AI Tutor} & ``That step is not correct. Here is the correction. Let $x$ be the fractional abundance of Co-60 and $1-x$ the abundance of Co-59. Solve $58.9901 = 58.9332(1-x) + 59.9338x$. Then find total cobalt atoms from 1.7516 g $\div$ 58.9901 g/mol $\times$ Avogadro's number and multiply by $x$ to obtain the number of Co-60 atoms.'' \newline
\textit{[Direct answer]} \\
\textbf{No-Direct-Answer AI Tutor} & ``Pause and test that idea before using it. If $x$ represents the fraction of Co-60, how would you express the Co-59 fraction? How can those two fractions and isotope masses produce the given apparent atomic mass?'' \newline
\textit{[Follow-up question]} \\
\textbf{Guided AI Tutor} & ``Actually, the reported atomic mass of the cobalt sample, 58.9901 amu, is a weighted average of the atomic masses of Co-59 and Co-60. This means it falls between the two values. How do you think we can use this information to find the percentage abundances of Co-59 and Co-60 in the sample?'' \newline
\textit{[Scripted Question; Address incorrect answer; Follow-up question]} \\
\bottomrule
\end{tabular}
\end{table*}

  \begin{figure}
      \centering
      \includegraphics[width=1\linewidth]{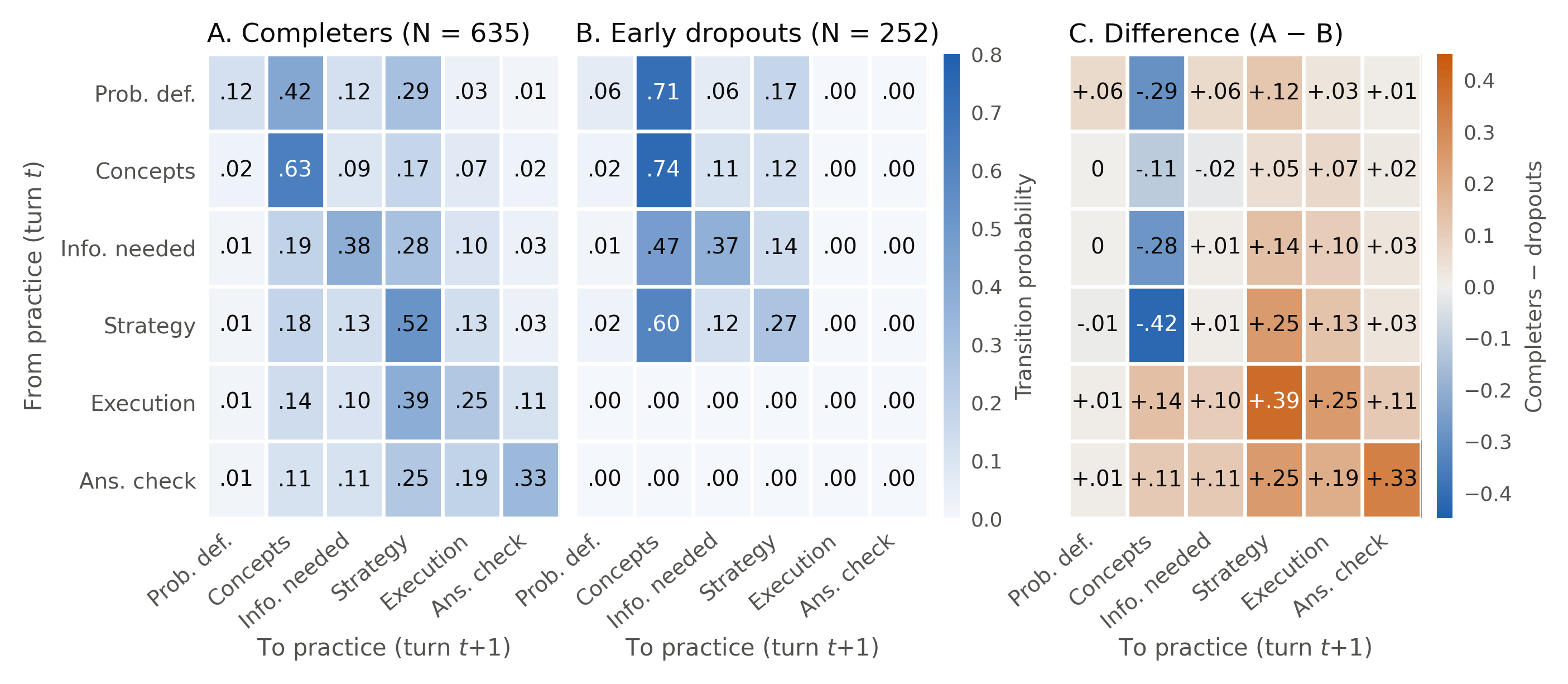}
      \caption{First-order practice-transition probabilities for (A) completers, (B) early dropouts, and (C) their difference $\Delta\mathbf{P}$. Transitions into execution and answer checking are zero for dropouts by definition and are not interpreted.}
      \label{fig:markov}
  \end{figure}
 
\subsection{RQ3: Early Dropouts Are Trapped in Concept Elicitation}
Figure~\ref{fig:markov} shows practice-transition matrices for completers and early dropouts. Overall, students who completed the problem showed a broader distribution of transitions and were generally less likely to transition back to relevant concepts and more likely to transition to strategy-focused practices. Both groups spent much of their time on relevant concepts, but completers moved forward through the problem-solving sequence: from problem definition, they moved either to relevant concepts ($P=.42$) or directly to strategy ($P=.29$); from strategy, they progressed to execution ($P=.13$); and from execution, to answer checking ($P=.11$).

Early dropouts showed a different structure before execution. From problem definition, 71\% of their transitions led to relevant concepts (vs.\ 42\% for completers). Once there, they stayed there: the concepts self-transition was $P=.74$. Even when they reached strategy, 60\% of their next turns returned to relevant concepts, compared with 18\% for completers ($\Delta=-.42$); the same backward pull appeared from information needed ($\Delta=-.28$). Because the tutor sets the practice of each exchange through its scripted sequence, these loops reflect the joint behavior of student and tutor: when students could not supply the concepts the tutor elicited, the tutor continued to elicit them. 
Combined with RQ1, where 52.3\% of impasses occurred in the concepts phase, this suggests that early attrition is associated with unresolved loops at the start of planning. Because dropouts are defined by never reaching execution, this finding describes where their sessions stalled rather than ruling out difficulties they might have met later.

\subsection{RQ4: The Value of Questioning Decays as Impasses Persist}
\paragraph{Impasse depth and state.} In Model~1, each additional turn at an impasse was associated with 13\% lower odds of recovery (AOR $=0.87$, 95\% CI [0.81, 0.94], $p<.001$), controlling for impasse type, practice, and tutor moves. Students expressing uncertainty (AOR $=1.37$, [1.17, 1.60]) or asking for help (AOR $=1.44$, [1.17, 1.77]) were more likely to recover than those making conceptual errors. Relative to relevant concepts, impasses during answer checking (AOR $=0.57$, [0.41, 0.80]) and strategy (AOR $=0.81$, [0.68, 0.96]) were less recoverable; other practices did not differ significantly.

\begin{table}[t]
\centering
\footnotesize
\setlength{\tabcolsep}{5pt}
\renewcommand{\arraystretch}{0.95}
\caption{Model 2: Recovery following selected two-turn scaffolding sequences ($M_{t-1}\rightarrow M_t$), relative to a first scripted question at onset, adjusted for depth, impasse type, and practice. Bold rows mark the reference and the two sequences compared in the text (repeating vs.\ switching after a scripted question).}
\label{tab:sequences}
\begin{tabular}{lrccc}
\toprule
\textbf{Sequence (turn $t-1 \rightarrow t$)} & \textbf{$n$} & \textbf{Recovery} & \textbf{AOR} & \textbf{$p$} \\
\midrule
\textbf{Onset $\rightarrow$ Scripted Q (ref.)} & 1,416 & \textbf{55.5\%} & -- & -- \\
Onset $\rightarrow$ Follow-up Q & 901 & 42.1\% & $0.71^{**}$ & .001 \\
Onset $\rightarrow$ Address incorrect & 676 & 39.5\% & $0.72^{**}$ & .006 \\
\midrule
\textbf{Scripted Q $\rightarrow$ Address incorrect} & 88 & \textbf{39.8\%} & \textbf{0.78} & \textbf{.291} \\
Scripted Q $\rightarrow$ Follow-up Q & 259 & 37.5\% & $0.69^{*}$ & .018 \\
Address incorrect $\rightarrow$ Address incorrect & 202 & 35.6\% & 0.81 & .245 \\
Follow-up Q $\rightarrow$ Follow-up Q & 372 & 33.1\% & $0.62^{**}$ & .003 \\
\textbf{Scripted Q $\rightarrow$ Scripted Q} & 299 & \textbf{28.1\%} & $\mathbf{0.48}^{***}$ & $\mathbf{<.001}$ \\
Partial answer $\rightarrow$ Partial answer & 56 & 19.6\% & $0.31^{**}$ & .002 \\
\midrule
Onset $\rightarrow$ Check understanding & 661 & 10.9\% & $0.11^{***}$ & $<.001$ \\
Address incorrect $\rightarrow$ Check understanding & 66 & 13.6\% & $0.19^{***}$ & $<.001$ \\
Follow-up Q $\rightarrow$ Check understanding & 126 & 7.9\% & $0.10^{***}$ & $<.001$ \\
Scripted Q $\rightarrow$ Check understanding & 73 & 8.2\% & $0.11^{***}$ & $<.001$ \\
\bottomrule
\multicolumn{5}{l}{\scriptsize $^{***}p<.001$, $^{**}p<.01$, $^{*}p<.05$. Standard errors clustered by session.}
\end{tabular}
\end{table}
\paragraph{Current and preceding moves.} In the tutor's response to the current impasse turn, scripted questions (AOR $=2.24$, [1.77, 2.85]), follow-up questions (AOR $=1.95$, [1.57, 2.44]), and addressing the incorrect answer (AOR $=1.36$, [1.11, 1.65]) were associated with higher recovery, whereas partial answers (AOR $=0.81$, [0.70, 0.94]) and checks for understanding (AOR $=0.37$, [0.28, 0.48]) were associated with lower recovery; direct answers and confirmation did not differ significantly. The latter result is likely partly a measurement artifact, because questions such as ``Does that make sense?'' invite brief affirmations that cannot be coded as completely correct. The tutor's preceding move also mattered: a scripted question on the preceding impasse turn was associated with lower recovery (AOR $=0.78$, [0.62, 0.98], $p=.032$), whereas preceding error-directed feedback was marginally associated with higher recovery (AOR $=1.26$, [0.99, 1.61], $p=.062$). Repeating the same dominant move had no additional effect (AOR $=0.99$). Full model estimates are provided in the supplementary material.

\paragraph{Scaffolding sequences.} Model~2 (Table~\ref{tab:sequences}) shows the same pattern at the level of sequences. A first scripted question at onset was followed by recovery in 55.5\% of cases. When a scripted question followed another scripted question, recovery fell to 28.1\%, halving the odds (AOR $=0.48$, [0.34, 0.67], $p<.001$). Moving from a scripted question to addressing the student's error preserved recovery (39.8\%; AOR $=0.78$, $p=.29$, not different from onset). Sequences ending in a check for understanding had the lowest recovery (7.9--13.6\%; AOR $0.10$--$0.19$, all $p<.001$). Model~2's dominant-move assignment rarely overrode the moves of interest: 90\% of tutor responses containing a scripted question were assigned that move. Recomputing the key sequences without any priority order, counting a sequence whenever both responses contained the move, reproduced the pattern (first scripted question: 54.4\%; scripted question $\rightarrow$ scripted question: 30.5\%; scripted question $\rightarrow$ address incorrect: 37.5\%). An order-free specification that added a (scripted question at $t-1$) $\times$ (scripted question at $t$) interaction to Model~1 likewise showed that a repeated scripted question carried substantially less benefit (AOR $=0.57$, 95\% CI [0.40, 0.82], $p=.002$).
 \begin{figure}
    \centering
    \includegraphics[width=\linewidth]{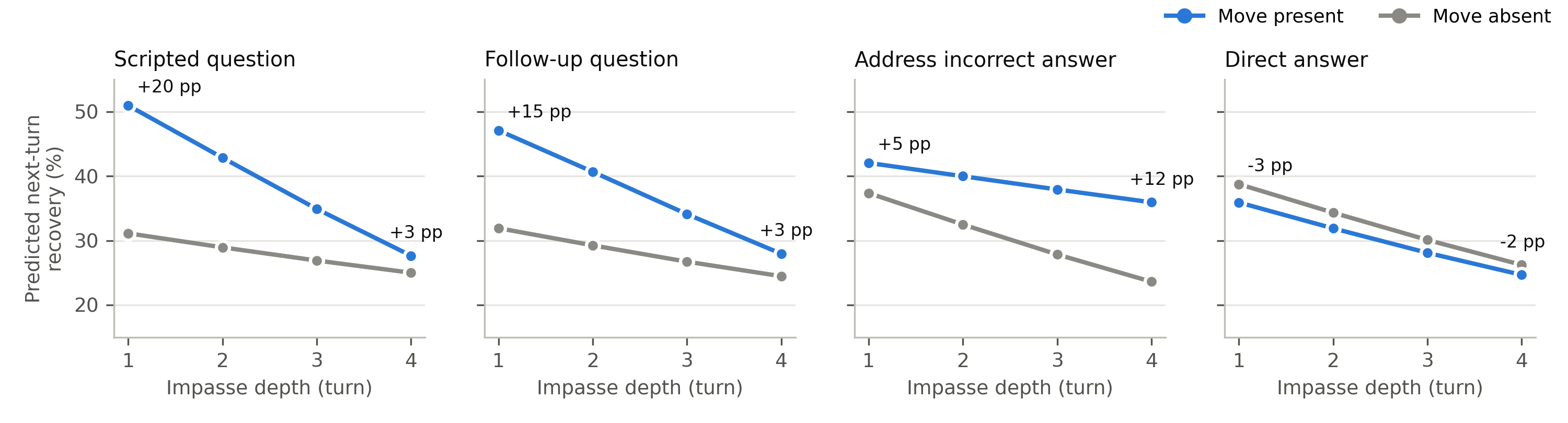}
    \caption{Average predicted next-turn recovery with each tutor move present versus absent, by impasse depth (Model~3). Labels give the difference in percentage points at depths 1 and 4.}
    \label{fig:interaction}
\end{figure}
\paragraph{Depth-dependent move effects.} Model~3 tests directly whether tutor move effects change as an impasse persists (Figure~\ref{fig:interaction}). The benefit of scripted questions (depth $\times$ scripted question AOR $=0.78$, [0.66, 0.91], $p=.002$) and of follow-up questions (AOR $=0.83$, [0.73, 0.95], $p=.008$) declined with each additional impasse turn, whereas the benefit of addressing the student's incorrect answer increased (AOR $=1.14$, [1.02, 1.28], $p=.026$). In predicted terms, a scripted question was associated with a 20 percentage-point (pp) higher recovery rate at onset (51.0\% vs.\ 31.1\%) but only 3 pp by the fourth impasse turn; follow-up questions fell from $+15$ to $+3$ pp; and addressing the error rose from $+5$ to $+12$ pp. Direct answers were associated with slightly lower recovery at every depth ($-3$ to $-2$ pp); their interaction with depth was not significant.

\subsubsection{Robustness}\label{sec:robust}
The depth-by-move interaction effects from Model~3 remain statistically significant even after accounting for clustering(scripted question $0.81$, $p=.006$; follow-up question $0.86$, $p=.028$; address incorrect $1.13$, $p=.030$). Because the continuous depth term assumes a constant per-turn decline, we also modeled depth categorically (1--5 and 6+, matching Figure~\ref{fig:depth_type}). Relative to onset, recovery odds were lower at every depth and lowest for the longest impasses (depth 2: $0.83$; depth 3: $0.61$; depth 4: $0.64$; depth 5: $0.55$; depth $\geq$6: $0.43$, 95\% CI [0.25, 0.74]; all $p\leq.01$).

\section{Discussion}\label{sec:discussion}

This study examined how a guardrailed LLM tutor responds as student impasses persist across turns. Across the controlled comparison and authentic dialogue analyses, a consistent distinction emerged between constraining tutor behavior and adapting assistance to an evolving interaction state. A more targeted tutoring protocol produced more varied tutor responses, but persistent impasses still exposed limits in how support changed after earlier interventions failed.

Our findings suggest that constraining an AI tutor and making it adaptive are distinct design problems. In the controlled comparison, simply instructing a tutor not to provide answers produced a relatively rigid questioning policy, whereas a more targeted tutoring protocol produced more varied responses across impasse contexts. However, even the deployed tutor, which was explicitly instructed to provide increasingly specific hints after incorrect or incomplete responses, sometimes continued with questioning after earlier questioning had failed to resolve the impasse. These findings suggest that pedagogical guardrails should specify not only constraints on undesirable behaviors such as answer revelation, but also how support should adapt when an initial intervention fails. This extends work on guardrailed LLM tutoring \cite{liffiton2023codehelp,bastani2025generative,jurenka2024towards} by shifting the design problem from response-level constraint to multi-turn adaptation, and similarly suggests that response-level evaluation criteria \cite{tack2022ai,maurya2025unifying} should be complemented by measures of whether tutor behavior changes appropriately across successive failed interactions. Our temporal analyses add a separate implication: the association between a tutor move and recovery is not fixed across an impasse. Questioning was strongly associated with recovery when an impasse first emerged, but this association weakened as the impasse persisted, whereas addressing an incorrect response became relatively more positively associated with recovery. This extends the assistance dilemma \cite{koedinger2007exploring} by suggesting that the appropriateness of support may depend not only on the learner's current difficulty, but also on how long that difficulty has persisted. It also qualifies work emphasizing the value of eliciting student reasoning \cite{vanlehn2003why,chi2001learning}: questioning was associated with higher recovery early in an impasse, but this association weakened following repeated unresolved turns. This pattern is consistent with contingent tutoring \cite{wood1999help} and graduated hinting in intelligent tutoring systems \cite{vanlehn2011relative}, where support becomes more explicit after failure. For LLM tutors, impasse persistence may therefore serve as a practical signal for when to consider shifting from elicitation toward more targeted assistance.

From a learning analytics perspective, impasses may be better represented as evolving interaction processes than as isolated incorrect turns. In our case, recovery declined as impasses deepened, and the associations between particular tutor moves and recovery changed with depth, indicating that the sequence and history of an interaction contain information that a single response cannot capture in isolation. This supports temporal approaches to learning analytics that treat the ordering and duration of events as analytically meaningful \cite{reimann2009time,chen2018critical}. Methodologically, impasse type, depth, and recovery provide relatively simple dialogue-derived measures for representing this process over time. For LLM-tutor evaluation, this shifts attention from whether an individual response is pedagogically appropriate toward whether assistance remains appropriate as the interaction unfolds. The trajectories of students who left sessions early further illustrate the value of this temporal view. These students became concentrated in repeated concept-elicitation loops before reaching problem execution, suggesting that stalled progress can emerge from the interaction between student difficulty and a tutor strategy that continues to elicit information the student is not providing. This resembles wheel-spinning \cite{beck2013wheel,gong2015towards,kai2018decision}, but at the level of tutoring dialogue rather than repeated skill practice. Repeated nonprogress may therefore provide learning analytics systems with an early signal of a stalled tutoring interaction, creating an opportunity to identify when adaptation may be needed before disengagement occurs. Taken together, these findings point toward graduated, state-sensitive assistance as a useful design target for LLM tutors. Rather than treating questioning or answer withholding as fixed pedagogical policies, tutors could use signals such as impasse type, persistence, and prior unsuccessful assistance to determine when more targeted support may be warranted. Such escalation need not imply immediately providing an answer; it could instead move from elicitation toward error-directed feedback, increasingly specific hints, or worked sub-steps as evidence accumulates that the current strategy is not resolving the impasse. This remains a design hypothesis rather than a demonstrated optimal policy, however, and should be tested through randomized comparisons that connect escalation strategies to both immediate recovery and subsequent learning.

\subsection{Limitations and Future Work}

Although the current work provides insights into students' impasse recovery process while working with an AI tutor, there are a few notable limitations. The authentic dialogue analyses are observational, so associations between tutor moves and recovery are not causal. Observations at greater impasse depths also represent the subset of episodes that remained unresolved at earlier turns, so depth-dependent associations may partly reflect differences in the composition of persistent impasses rather than changes caused by persistence itself. Recovery also captures only next-turn correctness, not longer-term learning. Our impasse categories reflect observable dialogue behaviors rather than latent cognitive or affective states, and the full corpus was coded with a human-validated LLM pipeline rather than exhaustive human annotation. Generalizability is also limited by the use of a single guided tutor in one undergraduate chemistry setting, while the controlled comparison examines tutor response policies rather than student outcomes. Future work should test these patterns across domains and tutor architectures, connect impasse recovery to longer-term learning, and experimentally compare escalation strategies.

\bibliography{references}


\end{document}